\pdfoutput=1

\documentclass[11pt]{article}

\usepackage[final]{acl}

\usepackage{times}
\usepackage{latexsym}

\usepackage[T1]{fontenc}

\usepackage[utf8]{inputenc}

\usepackage{microtype}

\usepackage{inconsolata}

\usepackage{graphicx}

\usepackage{array}
\usepackage{censor}
\usepackage{pdfpages}
\usepackage{appendix}
\usepackage{pdflscape}
\usepackage{textpos}
\usepackage{geometry}
\usepackage{makecell}
\usepackage{algorithm}
\usepackage{algpseudocode}
\usepackage{algorithmicx}
\usepackage{natbib}
\usepackage{float}

\title{Neural Morphological Tagging for Nguni Languages}

\author{Cael Marquard$^*$\,\,\,\,\,\, Simbarashe Mawere$^*$\,\,\,\,\,\, Francois Meyer\\
  Department of Computer Science \\
  University of Cape Town \\
  \texttt{\{mrqcae001,\,mwrsim003\}@myuct.ac.za, francois.meyer@uct.ac.za}}

\usepackage{times}
\usepackage{latexsym}

\usepackage{booktabs}
\usepackage{siunitx}
\usepackage{amsmath,amssymb,amsthm}
\usepackage{multirow}

\begin{document}
\maketitle
\def\thefootnote{*}\footnotetext{Equal contribution.}

\begin{abstract}
Morphological parsing is the task of decomposing words into morphemes, the smallest units of meaning in a language, and labelling their grammatical roles. It is a particularly challenging task for agglutinative languages, such as the Nguni languages of South Africa, which construct words by concatenating multiple morphemes. A morphological parsing system can be framed as a pipeline with two separate components, a segmenter followed by a tagger. This paper investigates the use of neural methods to build morphological taggers for the four Nguni languages. We compare two classes of approaches: training neural sequence labellers (LSTMs and neural CRFs) from scratch and finetuning pretrained language models. We compare performance across these two categories, as well as to a traditional rule-based morphological parser. Neural taggers comfortably outperform the rule-based baseline and models trained from scratch tend to outperform pretrained models. We also compare parsing results across different upstream segmenters and with varying linguistic input features. Our findings confirm the viability of employing neural taggers based on pre-existing morphological segmenters for the Nguni languages. 

\end{abstract}

\section{Introduction}

\begin{figure}[t]
    \centering
    \includegraphics[width=0.8\linewidth]{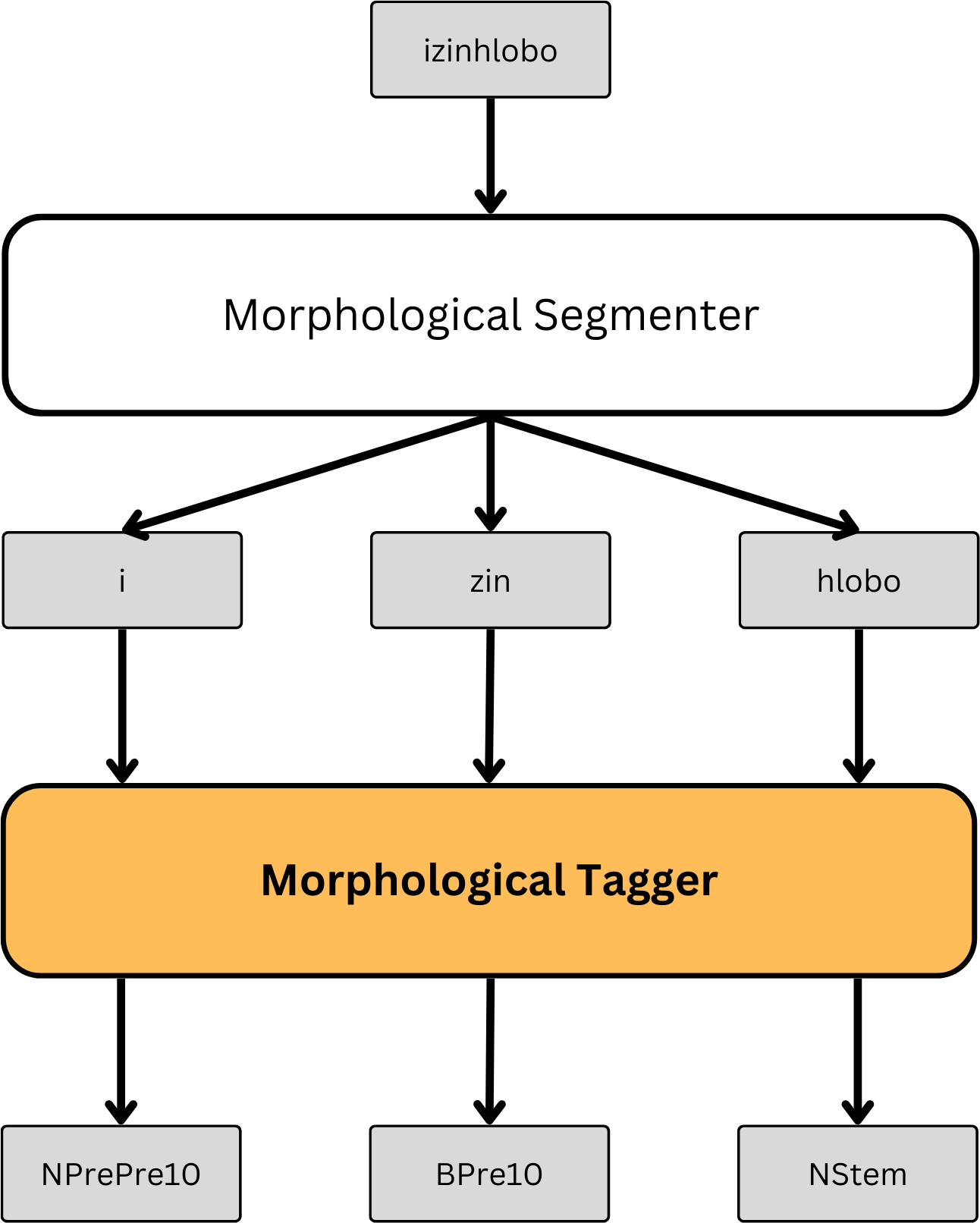}
    \caption{Morphological parsing as a two-step pipeline. We focus on tagging, training our taggers on the outputs of pre-existing morphological segmenters. }
    \label{fig:arch}
\end{figure}

The smallest unit of linguistic meaning that a word can be split into is known as a \textit{morpheme} \cite{matthews1991morphology}. Morphological parsing is the task of identifying the grammatical role of each morpheme within a word \cite{dutoit-puttkammer}. For example, ``izinhlobo'' (meaning ``types'' in isiZulu) is split into the morphemes ``i-zin-hlobo'', which is parsed as ``i[NPrePre10] - zin[BPre10] - hlobo[NStem]'' \cite{dataset-paper} (see \autoref{fig:arch}). Each bracketed tag labels the preceding morpheme with its grammatical function and noun class (if applicable). 


Morphological information is especially important for the Nguni languages, a group of related languages (isiNdebele, isiXhosa, isiZulu, and Siswati) spoken across South Africa by more than 23m home language speakers \citep{eberhard-etal-2019-ethnologue}. 
The Nguni languages are agglutinative, meaning that many words are created by aggregating multiple morphemes \cite{taljard_bosch_2006}. They are also written conjunctively---morphemes are concatenated into a single orthographic (space-delimited) word \cite{taljard_bosch_2006}. This can produce long, complex word forms consisting of several morphemes, such as the isiXhosa word ``andikambuzi'', which means ``I have not yet asked them'', composed of the morphemes ``a'', ``ndi'', ``ka'', ``m'', ``buza'', and ``i''.

As a result of this morphological complexity, morphological parsing is a challenging but important task for the Nguni languages. 
Despite this, few morphological parsers exist for the Nguni languages. Moreover, no existing parsers use neural methods, despite their established performance gains for linguistic annotation tasks \cite{plm-survey}. In this paper we explore the viability of neural morphological parsers for the Nguni languages.

Morphological parsing can be framed as a two-step pipeline \citep{parse-morph-rich, dutoit-puttkammer}, in which raw text is first segmented into morphemes, which are subsequently tagged with morphological labels. The first part of this pipeline is known as \textit{morphological segmentation}, while the second part is known as \textit{morphological tagging}. We visualise this pipeline for the isiZulu word ``izinhlobo'' in \autoref{fig:arch}. In this work we focus on the second subtask, morphological tagging. Instead of training models for the entire task, we make use of pre-existing morphological segmenters for the Nguni languages \citep{morph-segment-jan} and train neural taggers on top of their output. 

We train two classes of neural taggers -- neural sequence labellers trained from scratch and finetuned pretrained language models (PLMs). 
Our models trained from scratch are bi-LSTMs \citep{lstm-paper} and conditional random fields (CRFs) \cite{Lafferty2001ConditionalRF} with bi-LSTM features, using either morpheme or character-level input features. For PLMs, we finetune XLM-R-large \citep{conneau-etal-2020-unsupervised-xml-roberta}, Afro-XLMR-large \citep{alabi-etal-2022-afroxlmr}, and Nguni-XLMR-large \citep{meyer-etal-2024-ngunixlmr}, which respectively represent different levels of Nguni-language coverage.

We develop neural taggers based on two types of morphological segmentations: canonical and surface segmentations \citep{cotterell-etal-2016-joint}. Canonical segmentation decomposes a word into its constituent morphemes, in their standardised (pre-composed) form. For example, the isiXhosa word ``zobomi'' is canonically segmented into ``za-u-(bu)-bomi'', where some of the morphemes undergo spelling changes in word composition \cite{dataset-paper}. Surface segmentation decomposes a word into its constituent \textit{morphs}, which are the surface forms of morphemes as they appear in the composed word. For example, ``zobomi'' is surface-segmented into ``zo-bomi''. As demonstrated by this example, the canonical and surface-level segmentation of a word can differ.



We evaluate all our models in two settings. In the first, we test our taggers on the morphological segmentations available in our task dataset \citep{dataset-paper}. This provides an idealised setting in which we evaluate our models on gold-annotated segmentations, which we know to be correct, isolating tagging performance from segmentation mistakes. In the second setting, we test our taggers on the segmentations produced by the neural segmenters of \citet{morph-segment-jan}. These are model-predicted segmentations, so some segmentations will not align with morphological boundaries. This can lead to error propagation, in which segmentation errors degrade tagging performance. However, it also provides us with an estimate of how our taggers fare in a real-world setting in which the entire morphological parsing pipeline is predicted by neural models. 

Overall, we evaluate four variants of each model configuration -- trained on canonical/surface segmentations, and respectively tested on gold-annotated/model-predicted segmentations. Our study is an extensive investigation into the potential of neural parsers for all four Nguni languages. Our main findings can be summarised as follows:
\begin{itemize}
    \item Neural parsing comfortably outperforms our rule-based baseline, confirming the benefit of data-driven segmentation and tagging.
    \item Neural sequence labellers trained from scratch outperform finetuned PLMs on the morphological tagging subtask.
    \item With no access to gold-annotated morphological segmentations, canonical segmentations consistently leads to better parsing performance than surface segmentations.
\end{itemize}
We are the first to use neural models to train morphological taggers for the Nguni languages. To the best of our knowledge, our morphological parsing results represent state-of-the-art performance. Our models can be used to incorporate morphological information into downstream NLP models, which holds the potential to improve performance for the morphologically complex Nguni languages.



\section{Related Work}

Morphological parsing has been extensively studied in NLP \cite{parse-morph-rich, enhancing-with-morph-info}. Traditionally, it is performed by incorporating grammatical and morphological rules from the language into a finite-state transducer. This is a time-consuming process in which linguists construct hand-crafted rules \cite{procedure-morph-analysis-ancient}. As in other tasks of linguistic annotation \citep{plm-survey}, neural models provide an effective, data-driven solution approach to morphological parsing. 


Several works have trained a single model for morphological parsing, jointly modelling morphological segmentation and tagging \cite{seker-tsarfaty-2020-pointer, alecakir_2020, abudouwaili-etal-2023-joint, yshaayahu-levi-tsarfaty-2024-truly}. Alternatively, \citet{parse-morph-rich} propose a two-step architecture for parsing morphologically rich languages by first segmenting them into their morphemes and then tagging the morphemes with labels. 
Because morphological segmenters for Nguni languages already exist \cite{morph-segment-jan}, we choose to adopt this two-step pipeline approach, visualised in \autoref{fig:arch}. Despite the drawbacks of error propagation, training neural taggers alone is simpler than training joint segmentation-tagging models. The approach is also more modular, allowing for better segmenters to be substituted in as and when they are developed.  



A number of works have developed morphological segmenters, taggers, and parsers for the Nguni languages. 
ZulMorph \cite{zulMorph} is a rule-based canonical segmenter and tagger for isiZulu based on finite-state transducers. 
\citet{dutoit-puttkammer} develop data-driven (non-neural) canonical segmenters and taggers for all four Nguni languages. They apply TiMBL \cite{daelemans2004a}, a memory-based learning package, to the segmentation step, and MarMoT \cite{marmot, marmot-mueller}, a trainable CRF pipeline, to the tagging step. 
\citet{morph-segment-jan} were the first to apply neural methods to segmentation, using CRFs \cite{Lafferty2001ConditionalRF}, LSTMs \citep{lstm-paper}, and Transformers \citep{transformer-attention} to train canonical and surface-level segmenters for all four Nguni languages. They found that non-neural CRFs were best for surface segmentation, while Transformers outperformed the other methods in canonical segmentation.
Despite recent developments in neural models, such as sequence-to-sequence \cite{seq2seq_parsing} and sequence labeling models \cite{neural_labelling_crf}, no neural morphological taggers currently exist for the Nguni languages.

\section{Tagging Models} \label{method-prediction}

We now introduce our neural morphological taggers. 
Our models are trained on sequences of pre-segmented morphemes as input, and are tasked with assigning a morphological label to each morpheme. By focusing on the morphological tagging component of the morphological parsing pipeline (\autoref{fig:arch}), we can use established approaches to neural sequence tagging.

\subsection{Neural sequence labellers}

We train two types of neural models from scratch: bidirectional long short-term memory (bi-LSTM) networks \cite{lstm-paper} and conditional random fields (CRFs) \cite{Lafferty2001ConditionalRF} with bi-LSTM features.
Bi-LSTMs have previously been successfully applied to POS tagging \cite{Pannach_Meyer_Jembere_Dlamini_2022} and morphological segmentation \cite{morph-segment-jan} for the Nguni languages. 

CRFs are probabilistic models for sequence labelling. A CRF estimates the probability of a given output (label) sequence by modelling the interdependence of labels with each other, as well as their dependence on the input sequence. We use linear-chain CRFs because of their lower computational complexity (compared to higher-order CRFs). 
Traditionally, CRFs use a set of hand-crafted features to assign probabilities \cite{morph-segment-jan}. However, instead of designing these features by hand, a neural network can be used to automatically learn the features from the data \cite{morph-segment-jan, neural_ner_crf, neural_labelling_crf}. We choose a bi-LSTM to generate these features, as this has previously proved successful in POS tagging \cite{Pannach_Meyer_Jembere_Dlamini_2022} and morphological segmentation \cite{morph-segment-jan} for the Nguni languages. 

We experimented with several design choices for our neural models trained from scratch, varying the following factors:
\begin{itemize}
    \item \textbf{Feature level.} Models were trained on either morpheme-level or character-level input features, represented by learned embeddings in both cases. 
    For morpheme-level features, we replaced rare morphemes (<2 examples in the training data) with a special unknown token to help the model generalise to unseen data.
    For character-level features, we summed character embeddings to produce morpheme-level input embeddings.
    Surface models also have lowercase variants of these features.
    \item \textbf{Context level.} Models were trained on single words in isolation, or on entire sentences. Our goal was to investigate whether the additional context available to sentence-level sequence models would improve performance. 
\end{itemize}

\subsection{Pretrained language models}

We finetune the following three PLMs on our task:
\begin{enumerate}
    \item XLM-R-large \citep{conneau-etal-2020-unsupervised-xml-roberta}: a massively multilingual PLM trained on more than 100 languages, including isiXhosa.
    \item Afro-XLMR-large \citep{alabi-etal-2022-afroxlmr}: XLM-R further pretrained on 20 African languages, including isiXhosa and isiZulu.
    \item Nguni-XLMR-large \citep{meyer-etal-2024-ngunixlmr}: XLM-R adapted for the four Nguni languages.
\end{enumerate}

The models were selected to represent increasing levels of Nguni language pretraining coverage: XLM-R includes minimal Nguni data (only isiXhosa), Afro-XLMR adds isiZulu, while Nguni-XLMR specifically targets all four Nguni languages. We examine the degree to which these different levels of Nguni language inclusion influence downstream performance.


\section{Experimental Setup}

\subsection{Dataset}
We use the morphologically annotated dataset developed by \citet{dataset-paper}. It contains sentences from South African government publications, wherein each word is annotated with its morphological parse (segmentation and tags, as shown in \autoref{tab:corpus}), lemma, and part-of-speech. It contains 1,431 parallel paragraphs with roughly 50k words per language. The data is pre-split 90\%/10\% into train/test sets. The dataset contains only gold-standard canonical segmentations, so gold-standard surface segmentations were obtained through a script provided by \citet{morph-segment-jan}. Predicted segmentations for both canonical and surface forms were created by applying \citeauthor{morph-segment-jan}'s \cite{morph-segment-jan} to the raw text column of the dataset.

\begin{table}
    \centering
    \resizebox{\linewidth}{!}{
        \begin{tabular}{|c|c|c|}
             \hline
             \textbf{Word} & \textbf{Morphological analysis} \\
             \hline
             aliqela & a[RelConc6]-li[BPre5]-qela[NStem] \\
             kwibhunga & ku[LocPre]-i[NPrePre5]-(li)[BPre5]-bhunga[NStem]\\
             izincomo & i[NPrePre10]-zin[BPre10]-como[NStem] \\
             \hline
        \end{tabular}
    }
    \caption{Three examples from the isiXhosa part of the dataset used in our experiments \cite{dataset-paper}. Only the relevant aspects are included.}
    \label{tab:corpus}
\end{table}

\subsection{Model Configurations}

All our models are monolingually trained and evaluated on isiNdebele, isiXhosa, isiZulu, or Siswati. We evaluate four versions of each neural model, varying morphological input in the following ways.

\paragraph{Segmentation types} We train models for both types of morphological segmentation, allowing us to evaluate their respective difficulty. 

\begin{itemize}
  \item Canonical segmentation: decompose words into standardised morphemes (e.g., ``zobomi'' → ``za-u-(bu)-bomi'').
    \item Surface segmentation: decompose words into morphs as they appear in composed forms (e.g., ``zobomi'' → ``zo-bomi'').
\end{itemize}

\paragraph{Upstream segmentation} During testing, we assess performance across both idealised and practical scenarios.
\begin{itemize}
  \item Gold-annotated segmentations: apply taggers directly to the linguistically annotated, gold-standard morphological segmentations from the task dataset \citep{dataset-paper}. This provides an idealised setting in which morphological segmentations are known to be correct, isolating tagging performance from segmentation errors.

  \item Model-predicted segmentations: apply taggers to segmentations generated by neural segmenters \citep{morph-segment-jan}. We retrain their feature-based CRFs and Transformers on our training set to match our data setup. This simulates a real-world pipeline where segmentation is predicted, allowing for error propagation from segmentation to tagging. 
\end{itemize}


\subsection{Evaluation}
We use $F_1$ score to evaluate our models. 
We only evaluate morphological tagging performance, as opposed to full morphological parsing (segmentation + tagging). However, tagging inherently depends on segmentation in our setup, since models are trained on the pre-segmented morpheme sequences.

In our model-predicted segmentation setting, errors in predicted morphological segmentations can result in fewer or more predicted morphemes than morphological tags. As a result, in some instances we have to compute an $F_1$ score for predicted and target tag sequences of different lengths. We make use of the aligned multiset $F_1$ score proposed by \citet{aligned-f1}. This is an adaptation of the aligned segment $F_1$ score used in CoNLL18 \cite{conll-2018-conll}. The key difference is that the aligned multiset $F_1$ score bases token counts on the multiset intersection between the target and predicted sequences, so that target-prediction length mismatches are ameliorated.




We report both macro $F_1$ and micro $F_1$ in our results. 
Micro $F_1$ is a calculated by counting the number of true positives/negatives and false positives/negatives for all classes. More common tags therefore have a greater effect on the Micro $F_1$ score. With one tag per item, it is equivalent to accuracy. 
Macro $F_1$ calculates the per-class $F_1$ score and averages them, weighting all tags equally irrespective of frequency. 
A high macro $F_1$ score indicates good performance across all tags, including rare tag types. 
We focused on macro $F_1$ during hyperparameter tuning and in discussing our results, as we consider it important for our models to perform well on rare tags. Our evaluation dataset \citep{dataset-paper} is imbalanced from a tag perspective, so macro $F_1$ is the more challenging metric to optimise than micro $F_1$.

\subsection{Hyperparameters}

The morphologically annotated dateset \citep{dataset-paper} is split into train and test sets, but does not include a validation set. To prevent over-fitting hyperparameters to the test set, we created our own held-out validation set from 10\% of the training set. Hyperparameter settings were tuned to maximise macro $F_1$ scores on the validation dataset.

\begin{table}[t]
    \small
        \centering
        \begin{tabular}{lc}
             \toprule
             \textbf{Hyperparameter} & \textbf{Search space} \\
             \midrule
             \multicolumn{2}{c}{\textbf{Neural sequence labellers}} \\
             \midrule
             Learning rate & $[10^{-6}, 10^{-1}]$ \\
             Weight decay & $\{0\} \cup [10^{-10}, 10^{-3}]$ \\
             Hidden state size &$\{2^x : 6 \leq x \leq 11\}$\\
             Dropout & $\{0, 0.1, 0.2, 0.3\}$\\
             Gradient clip & $\{0.5, 1, 2, 4, \infty\}$\\
              \midrule
            \multicolumn{2}{c}{\textbf{Finetuned PLMs}} \\
             \midrule
             Learning rate & $\{10^{-5}, 3\times10^{-5}, 5\times10^{-5}\}$\\ 
             Epochs & $\{5, 10, 15\}$\\
             Batch size & $\{8, 16, 32\}$\\
             \bottomrule
        \end{tabular}
        \caption{The hyperparameter ranges of our grid search.}
        \label{tab:hyperparam-scratch}
    \end{table}

For our models trained from scratch, we performed a grid search over the hyperparameter ranges shown in \autoref{tab:hyperparam-scratch}. We tuned our hyperparameter settings on isiZulu only, because including other languages would lead to a computationally infeasible grid search. Once the best parameters for isiZulu were found, these configurations were applied to the other languages. 
For our PLMs, we also performed a grid search over finetuning hyperparameters over the grid shown in \autoref{tab:hyperparam-scratch}. 


After we settled on our final hyperparameter settings based on validation set performance, we retrained models on the full, original training set (including our newly created validation set) and evaluated them on the test set. For each model configuration, we train/finetune five models with different random seeds and report the average evaluation metrics.

\subsection{Baselines}

We compare our neural methods to ZulMorph \cite{zulMorph}, a traditional, rule-based parser for IsiZulu. ZulMorph is based on finite-state transducers with manually incorporated grammatical rules, stems, and affixes for isiZulu. We use the ZulMorph demo \cite{pretorius2018a} to evaluate its performance on the test set. Since ZulMorph both segments and tags the input data, we compare it to our taggers trained on model-predicted segmentations.

\section{Results}

The results based on gold-annotated segmentations are shown in \autoref{tab:canon-res}, while those based on model-predicted segmentations are shown in \autoref{tab:end-to-end}.

Overall, our results demonstrate the effectiveness of neural models on the challenging task of morphological tagging for Nguni languages. Our best-performing models based on gold-annotated canonical segmentations consistently achieve micro $F_1$ scores above 90\% and macro $F_1$ scores above 60\%. Even without access to the gold morphological annotations, with models tested on the predicted canonical segmentations of \citet{morph-segment-jan}, our best models consistently achieve micro $F_1$ scores above 80\% and macro $F_1$ scores above 55\%. 
This confirms the feasibility of basing the full morphological parsing pipeline on neural models.



\begin{table*}[t]
\centering \small
    \begin{tabular}{ lc cc cc cc c }

    \toprule
    \textbf{Model} & \multicolumn{2}{c}{\textbf{IsiZulu}} & \multicolumn{2}{c}{\textbf{IsiNdebele}} & \multicolumn{2}{c}{\textbf{IsiXhosa}} & \multicolumn{2}{c}{\textbf{Siswati}} \\ 
    \midrule
    & Mac $F_1$ & Mic $F_1$ & Mac $F_1$ & Mic $F_1$ & Mac $F_1$ & Mic $F_1$ & Mac $F_1$ & Mic $F_1$ \\

    \midrule
    \multicolumn{9}{c}{\textbf{Canonical segmentations as annotated in \citet{dataset-paper}}} \\ 
    \midrule
    
    \textbf{Trained from scratch} & & & & & & & & \\
    \midrule
    \textbf{Word-level} & & & & & & & & \\
    bi-LSTM, character-sum & \underline{\textbf{66.9}} & \underline{\textbf{92.4}} & 67.2 & \underline{\textbf{91.9}} & 72.3 & 95.2 & 66.5 & 91.2\\
    bi-LSTM, morpheme & 66.6 & 92.1 & 67.7 & 91.8 & 71.5 & 94.9 & 65.5 & 91.0\\
    \textbf{Sentence-level} & & & & & & & & \\
    bi-LSTM, character-sum & 64.6 & 91.6 & 66.6 & 91.0 & 72.1 & 95.5 & 64.7 & 90.8\\
    bi-LSTM, morpheme & 66.0 & 92.1 & 67.9 & 91.6 & 74.7 & 95.7 & \underline{\textbf{67.2}} & 91.3\\
    CRF, character-sum & 65.7 & 92.1 & 67.3 & 91.4 & 74.7 & \underline{\textbf{95.9}} & 66.0 & 91.4\\
    CRF, morpheme & 66.1 & 92.3 & \underline{\textbf{68.1}} & 91.6 & \underline{\textbf{75.3}} & 95.8 & 67.2 & \underline{\textbf{91.4}}\\
    \midrule
    \textbf{Pretrained language models} & & & & & & & & \\
    \midrule
    \textbf{Word-level} & & & & & & & & \\
    Afro-XLMR & \textbf{62.5} & \textbf{92.0} & 62.3 & 91.4 & 67.9 & 95.1 & \textbf{63.3} & \textbf{91.3}\\
    Nguni-XLMR & 61.9 & 92.0 & 62.8 & 91.5 & \textbf{68.1} & \textbf{95.1} & 61.8 & 90.7\\
    XLM-R-large & 61.8 & 91.8 & \textbf{63.6} & \textbf{91.6} & 67.4 & 95.0 & 62.9 & 91.2\\

    \midrule
    \multicolumn{9}{c}{
    \textbf{Surface segmentations extrapolated from \citet{dataset-paper} by script from \citet{morph-segment-jan}}
    } \\ 
    \midrule

    \textbf{Trained from scratch} & & & & & & & & \\
    \midrule
    \textbf{Sentence-level} & & & & & & & & \\
    bi-LSTM, character-sum & 63.3 & 90.7 & 65.2 & 90.4 & 73.6 & 94.7 & 61.3 & 89.6\\
    bi-LSTM, character-sum-lower & 63.2 & 90.8 & 65.4 & 90.4 & 73.7 & 94.7 & 60.8 & 89.7\\
    bi-LSTM, morpheme & 65.6 & 91.3 & 68.4 & 91.1 & \underline{\textbf{76.1}} & 95.1 & \underline{\textbf{65.9}} & 90.6\\
    bi-LSTM, morpheme-lower & \underline{\textbf{66.0}} & \underline{\textbf{91.3}} & \underline{\textbf{68.7}} & \underline{\textbf{91.2}} & 76.0 & \underline{\textbf{95.3}} & 65.8 & \underline{\textbf{90.7}}\\
    \midrule
    \textbf{Pretrained language models} & & & & & & & & \\
    \midrule
    \textbf{Word-level} & & & & & & & & \\
    Afro-XLMR & 43.8 & 72.8 & 47.7 & 77.4 & 52.3 & 78.5 & 23.4 & 55.6\\
    Nguni-XLMR & \textbf{44.1} & \textbf{73.1} & \textbf{48.1} & \textbf{77.5} & \textbf{52.4} & \textbf{79.0} & \textbf{23.9} & \textbf{56.6}\\
    XLM-R-large & 43.1 & 72.6 & 48.0 & 77.5 & 51.7 & 78.1 & 22.7 & 55.4\\
    
    \bottomrule

    \end{tabular}
     \caption{Results for models evaluated on gold-annotated segmentations, given as percentages. This provides an idealised training setting in which all morphological segmentations are correct, allowing us to isolate the performance of morphological tagging. The best models for each approach (pretrained or from scratch) is \textbf{bolded}, while the best for each segmentation type (surface or canonical) is \underline{underlined}.}
    \label{tab:canon-res}
\end{table*}

\begin{table*}[t]
\centering \small
    \begin{tabular}{ lc cc cc cc c }

    \toprule
    \textbf{Model} & \multicolumn{2}{c}{\textbf{IsiZulu}} & \multicolumn{2}{c}{\textbf{IsiNdebele}} & \multicolumn{2}{c}{\textbf{IsiXhosa}} & \multicolumn{2}{c}{\textbf{Siswati}} \\ 
    \midrule
    & Mac $F_1$ & Mic $F_1$ & Mac $F_1$ & Mic $F_1$ & Mac $F_1$ & Mic $F_1$ & Mac $F_1$ & Mic $F_1$ \\

    \midrule
    \multicolumn{9}{c}{\textbf{ZulMorph online demo \cite{pretorius2018a}}} \\ 
    \midrule

    ZulMorph & 34.0 & 71.8 & & & & & & \\
    
    \midrule
    \multicolumn{9}{c}{\textbf{Canonical segmentations as predicted by \citet{morph-segment-jan}}} \\ 
    \midrule
    \textbf{Trained from scratch} & & & & & & & & \\
    \midrule
    \textbf{Word-level} & & & & & & & & \\
    bi-LSTM, character-sum & \underline{\textbf{60.0}} & \underline{\textbf{85.8}} & 57.8 & \underline{\textbf{84.1}} & 67.9 & 92.3 & 57.0 & 85.0\\
    bi-LSTM, morpheme & 58.3 & 85.5 & 58.3 & 84.1 & 67.0 & 92.2 & 55.7 & 84.7\\
    \textbf{Sentence-level} & & & & & & & & \\
    bi-LSTM, character-sum & 57.5 & 85.1 & 57.3 & 83.4 & 68.1 & 92.7 & 55.5 & 84.8\\
    bi-LSTM, morpheme & 58.4 & 85.7 & 58.3 & 83.8 & 70.7 & 93.0 & 57.3 & 85.2\\
    CRF, character-sum & 58.1 & 85.5 & 58.4 & 83.8 & 69.8 & \underline{\textbf{93.1}} & 57.2 & \underline{\textbf{85.4}} \\
    CRF, morpheme & 58.7 & 85.7 & \underline{\textbf{58.5}} & 83.7 & \underline{\textbf{71.1}} & 93.1 & \underline{\textbf{57.8}} & 85.3\\
    \midrule
    \textbf{Pretrained language models} & & & & & & & & \\
    \midrule
    \textbf{Word-level} & & & & & & & & \\
    Afro-XLMR & \textbf{55.3} & 85.5 & 54.6 & 84.0 & 63.4 & 92.4 & \textbf{53.4} & \textbf{85.1}\\
    Nguni-XLMR & 54.8 & \textbf{85.5} & 54.5 & 83.9 & \textbf{64.4} & \textbf{92.6} & 52.5 & 84.6\\
    XLM-R-large & 54.4 & 85.4 & \textbf{55.4} & \underline{\textbf{84.1}} & 63.5 & 92.5 & 52.9 & 85.0\\
    \midrule
    \multicolumn{9}{c}{\textbf{Surface segmentations as predicted by \citet{morph-segment-jan}}}\\ 
    
    \midrule
    \textbf{Trained from scratch} & & & & & & & & \\
    \midrule
    \textbf{Sentence-level} & & & & & & & & \\
    bi-LSTM, character-sum & 53.6 & 79.6 & 52.8 & 78.3 & 65.6 & \underline{\textbf{87.7}} & 51.8 & 80.4\\
    bi-LSTM, character-sum-lower & 53.3 & 79.6 & 52.9 & 78.2 & 65.2 & 87.5 & 51.6 & 80.4\\
    bi-LSTM, morpheme & 55.0 & 79.7 & \underline{\textbf{54.7}} & 78.4 & 68.0 & 87.4 & 55.2 & 81.0\\
    bi-LSTM, morpheme-lower & \underline{\textbf{55.3}} & \underline{\textbf{79.7}} & 54.6 & \underline{\textbf{78.5}} & \underline{\textbf{68.2}} & 87.6 & \underline{\textbf{55.8}} & \underline{\textbf{81.0}}\\
    \midrule
    \textbf{Pretrained language models} & & & & & & & & \\
    \midrule
    \textbf{Word-level} & & & & & & & & \\
    Afro-XLMR & 43.6 & 72.8 & 46.9 & 77.4 & \textbf{51.9} & 78.5 & 23.0 & 55.7\\
    Nguni-XLMR & \textbf{43.9} & \textbf{73.0} & 46.9 & 77.4 & 51.7 & \textbf{78.8} & \textbf{23.7} & \textbf{56.3}\\
    XLM-R-large & 43.1 & 72.7 & \textbf{47.7} & \textbf{77.5} & 51.4 & 78.0 & 22.1 & 55.4\\
    \bottomrule
    
    \end{tabular}
    \caption{Results for models evaluated on model-predicted segmentations, given as percentages. This evaluates the combined use of neural methods for segmentation and tagging, without access to morphological annotations. The best models for each approach (pretrained or from scratch) is \textbf{bolded}, while the best for each segmentation type (surface or canonical) is \underline{underlined}.}
    \label{tab:end-to-end}
\end{table*}


\paragraph{Comparison to rule-based parsing} The neural models comfortably outperform our rule-based baseline, ZulMorph, on isiZulu morphological tagging. ZulMorph \cite{pretorius2018a} achieves a macro $F_1$ of 34\% and micro $F_1$ of 71.8\% on the test-set. All our isiZulu models surpass this performance, ranging from macro $F_1$s of 43.1\% to 60\% and micro $F_1$s of 72.7\% to 85.8\%. 



Since ZulMorph is rule-based and contains manually-incorporated stems and affixes, it likely struggles to generalise to unseen data. For instance, ZulMorph failed to segment and parse ``wezentuthuko'', and instead produced ``wezentuthuko	+?''. Conversely, the neural models do not explicitly incorporate any information. The models are able to classify text even when there are unknown morphemes present in the text, based on the surrounding context of known morphemes.

\paragraph{Macro vs Micro $F_1$} Macro $F_1$ is consistently lower than corresponding micro $F_1$ scores. This highlights one of the difficulties of morphological tagging for the Nguni languages. The tag set is large and unevenly distributed in the dataset, which make it challenging to accurately model rare tags. This imbalance would explain the mismatch between macro and micro $F_1$ for neural models, since they are not adequately exposed to rare tags during training. However, the mismatch persists for ZulMorph \cite{zulMorph}  (see \autoref{tab:end-to-end}), which is based on gramatically informed rules, as opposed to being data-driven. This could indicate that some tags are inherently harder to disambiguate.


\subsection{Training neural taggers from scratch}


As shown in Tables \ref{tab:canon-res} and \ref{tab:end-to-end}, sentence-level models trained from scratch tended to outperform their word-level counterparts. Sentence-level models are trained on the entire sentence as context, which may allow them to use grammatical dependencies to improve tagging. For example, in the isiXhosa sentence ``ipolisa liyahamba'', the word ``ipolisa'' is in noun class 5. The shorted prefix ``i'' (``\textbf{i}polisa'') is ambiguous and also appears in class 9 nouns, such as ``iteksi''. However, combining it with the subject concord for class 5 ``li'' (``\textbf{li}yahamba'') provides the information required to correctly disambiguate and tag ``ipolisa'' as class 5.

Morpheme-level embeddings outperformed character-summing embeddings. While one might expect character-level modelling to improve generalisation across morphemes, this is not necessarily the case. Morphemes representations have previously been shown to be highly effective for syntactic tasks \cite{ustun-etal-2018-characters}. 
For our task, morpheme-level embeddings allow the model to be more sensitive to small changes in morphemes. For example, the morphemes ``ng'' and ``nga'' differ by a single character, but can have totally different meanings (``ng'' can be a copulative prefix and ``nga'' can be an adverb prefix). With character-summed representations, the two morphemes will have highly similar embeddings. With morpheme-level embeddings, each morpheme embedding is learned separately. For rare or previously unseen morphemes, the morpheme-level model is forced to rely on contextual grammatical information (within the word or surrounding sentence), which provides a more reliable grammatical signal than the number of overlapping characters between morphemes.


We do not find substantial performance differences between bi-LSTMs and bi-LSTM CRFs. This indicates that explicitly modeling grammar through tag dependence presents limited advantage. Bi-LSTMs are able to encode such grammatical dependencies, based on morpheme co-occurrence patterns, in their hidden representations. 

\subsection{Pretrained language models} \label{results-fine-tuning}
As shown in Tables \ref{tab:canon-res} and \ref{tab:end-to-end}, training models from scratch outperformed finetuning PLMs. This contrasts with previous work on linguistic annotation tasks, in which pretrained solutions have outperformed models trained from scratch \cite{plm-survey, alabi-etal-2022-afroxlmr}. However, it does align with related work for the Nguni languages, which have achieved high performance levels with neural models trained from scratch \cite{morph-segment-jan, Pannach_Meyer_Jembere_Dlamini_2022}.

Due to computational constraints, we did not finetune PLMs on sentence-level input. The pretrained contextual representations of PLMs are well suited to take advantage of sentence-level context, so it is possible that finetuning sentence-level versions of our PLMs could improve their performance. We leave the exploration of sentence-level PLMs for Nguni-language morphological tagging to future work.

Another factor which could contribute to PLM performance degradation is subword tokenisation. While our models trained from scratch use character or morpheme-level representations, our PLMs are constrained to finetune representations for the subword tokens produced by their pretrained tokenisers. In pretraining, the tokeniser segments raw words. In finetuning, the tokeniser segments pre-segmented morphemes. This misalignment could impede the model's ability to leverage pretrained knowledge during finetuning, since the subword tokens learned in pretraining do not match those of finetuning. This also leads to irregular, morphologically unsound subword tokens. For example, the XLM-R SentencePiece tokeniser \cite{conneau-etal-2020-unsupervised-xml-roberta, sentencepiece} segments, which is the tokeniser for all our PLMs, segments the isiXhosa morpheme ``-bandela'' into ``-ba'', ``\#ndel'', ``\#a'', which is morphologically meaningless. In our pipeline setup for morphological parsing, it is not obvious how to bridge the mismatch between pretraining and finetuning subword tokenisation. It should be viewed as a limitation of PLMs. With neural models trained from scratch, we have the freedom to design our own morphological input features.

\subsection{Models based on surface segmentations}

In both Tables \ref{tab:canon-res} and \ref{tab:end-to-end}, the top half of each table reports results for models trained on canonical segmentations (morphemes), while the bottom half reports results for surface-level segmentations (morphs). In general, canonically-based tagging scores are higher than surface-level tagging. The performance gap is particularly notable and consistent for models trained on model-predicted segmentations. While canonical and surface-level tagging scores cannot be directly compared (for some words, the tag sequence will not be the same), our results clearly show that training taggers on top of canonical segmenters is more effective than doing so with surface-level segmenters. We attribute this to two factors.



Firstly, the surface segmentation of a word provides less grammatical information to models than the canonical segmentation. For instance, the word ``kwicandelo'' is canonically segmented as ``ku-i-(li)-candelo'' and surface segmented as ``kw-i-candelo'' \cite{dataset-paper}. Critically, the ``(li)'' morpheme is lost, which is part of the noun prefix for class 5. The only morpheme left for the noun prefix is thus ``i''. However, this on its own is ambiguous, and could be the noun prefix for class 5 or class 9. In this case, the canonical tagger would have more information relevant to the tagging decision than the surface tagger.

Secondly, there is often a length mismatch between the surface and canonical morphemes in a word. For example, ``kubomi'' is canonically segmented into ``ku-u-(bu)-bomi'', but surface-segmented into ``ku-bomi''. We evaluate our model on gold-annotated data, which include morphological tags for each word. In a case like ``kubomi'', this would limit performance to 50\% accuracy in the best case scenario. In general, this length mismatch limits the performance of models based surface-level segmentations.

\section{Conclusion}

In this paper, we explored the feasibility of neural morphological taggers for the Nguni languages. We divide morphological parsing into two subtasks, segmentation and tagging, focussing on the latter. We investigate bi-LSTMs and CRFs trained from scratch, as well as finetuned PLMs. 
Our neural models comfortably outperform a rule-based baseline, while our models trained from scratch outperform PLMs. Models based on canonical segmentations outperform their surface-level counterparts.

We identify several promising directions for future research to build on our findings. Firstly, our PLM taggers could potentially be improved, either by finetuning on sentence-level input or by exploring ways to align the mismatch between subword tokenisation in pretraining and finetuning. Furthermore, our parsers can be used to incorporate morphological information into downstream task models \cite{enhancing-with-morph-info}. This has been shown to improve performance in tasks such as language modelling \citep{nzeyimana-niyongabo-rubungo-2022-kinyabert} and machine translation \citep{nzeyimana-2024-low}, but has not been explored for the Nguni languages.

\section*{Limitations} 

Our study is limited to the Nguni languages, so our findings may not generalise to other language families or typologies like the Sotho-Tswana languages whose morphology is disjunctive. Further experimentation is needed to validate whether training taggers on model-predicted morphological segmentations is viable for languages with different morphological structures. 
That being said, the promising performance of our models on the Nguni languages suggests that similar neural approaches could be beneficial for other low-resource, morphologically complex languages.


Additionally, while our models trained from scratch consistently outperformed finetuned PLMs, we do not definitively conclude that PLMs are inferior for this task. As discussed in \autoref{results-fine-tuning}, because of computational constraints we did not test sentence-level PLMs. Incoporating sentence-level context could improve PLM performance to be competitive with models trained from scratch. 
We would need to run further experiments with sentence-level finetuning to evaluate the full potential of PLMs for this task. 
\bibliography{custom, anthology_0, anthology_1}

\appendix

\end{document}